\documentclass[lettersize,journal]{styley}
\usepackage{amsmath,amsfonts}
\usepackage{algorithmic}
\usepackage{algorithm}
\usepackage{array}
\usepackage[caption=false,font=normalsize,labelfont=sf,textfont=sf]{subfig}
\usepackage{textcomp}
\usepackage{stfloats}
\usepackage{url}
\usepackage{verbatim}
\usepackage{graphicx}
\usepackage{cite}

\usepackage{bbding} 
\usepackage{threeparttable}
\usepackage{float} 
\usepackage{booktabs}
\usepackage{multirow}

\begin{document}

\title{LLMI3D: MLLM-based 3D Perception from a Single 2D Image}

\author{Fan Yang, Sicheng Zhao, Yanhao Zhang, Hui Chen, Haonan Lu, Jungong Han, Guiguang Ding
\thanks{
Fan Yang, Sicheng Zhao, Hui Chen, and Guiguang Ding are with BNRist, Tsinghua University, Beijing, 100084, China.
Fan Yang and Guiguang Ding are also with the School of Software, Tsinghua University, Beijing, 100084, China.
(e-mails: yfthu@outlook.com, schzhao@gmail.com, \quad \quad jichenhui2012@gmail.com, dinggg@tsinghua.edu.cn)
}
\thanks{Jungong Han is with the Department of Automation, Tsinghua University, Beijing, 100084, China. (e-mails: jungonghan77@gmail.com)}
\thanks{Yanhao Zhang and Haonan Lu are with OPPO AI Center. 
(e-mails: zhangyanhao@oppo.com, luhaonan@oppo.com)}
\thanks{Corresponding Authors: Sicheng Zhao and Guiguang Ding.}
}

\maketitle

\def\abb{LLMI3D}
\def\abbdata{IG3D}

\begin{figure*}[!h]
\centering
\includegraphics[width=\linewidth]{Figures/coverfigure_02112153.pdf}
\caption{Our \abb~endows MLLMs with 3D perception capabilities. When provided with a question or description, our \abb~can return the object of interest and its 3D bounding box (bbox) in 3D space. Across various datasets, our \abb~significantly outperforms existing methods.}
\label{coverfigure}
\end{figure*}

\begin{abstract}
Recent advancements in autonomous driving, augmented reality, robotics, and embodied intelligence have necessitated 3D perception algorithms. 
However, current 3D perception methods, especially specialized small models, exhibit poor generalization in open scenarios.
On the other hand, multimodal large language models (MLLMs) excel in general capacity but underperform in 3D tasks, due to weak 3D local spatial object perception, poor text-based geometric numerical output, and inability to handle camera focal variations.
To address these challenges, we propose the following solutions: Spatial-Enhanced Local Feature Mining for better spatial feature extraction, 3D Query Token-Derived Info Decoding for precise geometric regression, and Geometry Projection-Based 3D Reasoning for handling camera focal length variations. We employ parameter-efficient fine-tuning for a pre-trained MLLM and develop \abb, a powerful 3D perception MLLM. Additionally, we have constructed the \abbdata~dataset, which provides fine-grained descriptions and question-answer annotations. Extensive experiments demonstrate that our \abb~achieves state-of-the-art performance, outperforming other methods by a large margin. 
\end{abstract}

\section{Introduction}

With the rapid development of deep learning, 2D perception tasks such as object detection, instance segmentation, and visual grounding have achieved remarkable progress \cite{li2024lmeye,zhang2023one,detr,zhao2024multi}. However, the real world is three-dimensional, and many practical applications, such as autonomous driving, robotics, augmented reality, and embodied intelligence, demand enhanced spatial perception.
Traditional 2D methods can no longer meet these demands. Therefore, researchers have introduced the concept of three-dimensional perception, which involves inferring the location, dimension, and pose of objects in three-dimensional space to achieve accurate predictions of their spatial positions \cite{an2023sp,shen2022depth,hua2022weakly,DBLP:conf/cvpr/MousavianAFK17,DBLP:conf/aaai/QinWL19}.

Most 3D perception techniques utilize LiDAR point clouds \cite{DBLP:conf/cvpr/LangVCZYB19} or camera images. LiDAR offers excellent depth prediction but is expensive and complex. Image-based methods \cite{sun2023learning}, being more affordable and easily integrable, are widely used in autonomous driving and robotics.

In recent years, various specialized models have been developed for image-based 3D perception. However, they face several limitations: 
1. Single-modal 3D detection models lack the ability to precisely locate a specified interested object based on textual input, limiting their effectiveness in following user instructions.
2. These models possess limited logical reasoning and question-answering capabilities due to insufficient real-world knowledge and common sense.
3. They demonstrate weak generalization in open and cross-domain settings, being restricted to predefined categories and scenes within the training dataset. 
\begin{table*}[]
\centering
\resizebox{1.0\hsize}{!}{
\begin{threeparttable}
\caption{Existing specialized small models and multimodal large language models have various limitations. Only our approach, \abb, exhibits comprehensive and robust 3D perception capabilities.}
\label{introcampare}
\begin{tabular}{@{}ccccccccc@{}}
\toprule
Methods & LLM & Instruction Understanding & Logical Reasoning & Question Answering & Open Vocabulary & Local 3D Feature Extracting & Focal Length Variation Handling & 3D Box Outputting \\ \midrule
MonoDETR~\cite{monodetr} & \XSolidBrush & \XSolidBrush & \XSolidBrush & \XSolidBrush & \XSolidBrush & \Checkmark & \XSolidBrush & \Checkmark \\
Omni3D~\cite{omni3d} & \XSolidBrush & \XSolidBrush & \XSolidBrush & \XSolidBrush & \XSolidBrush & \Checkmark & \Checkmark & \Checkmark \\
Mono3DVG~\cite{mono3dvg} & \XSolidBrush & \Checkmark & \XSolidBrush & \XSolidBrush & \XSolidBrush & \Checkmark & \XSolidBrush & \Checkmark \\
DeepSeek-VL2~\cite{wu2024deepseek} & \Checkmark & \Checkmark & \Checkmark & \Checkmark & \Checkmark & \XSolidBrush & \XSolidBrush & \XSolidBrush \\
InternVL2.5~\cite{chen2024expanding} & \Checkmark & \Checkmark & \Checkmark & \Checkmark & \Checkmark & \XSolidBrush & \XSolidBrush & \XSolidBrush \\
\abb~(Ours) & \Checkmark & \Checkmark & \Checkmark & \Checkmark & \Checkmark & \Checkmark & \Checkmark & \Checkmark \\ \bottomrule
\end{tabular}
\end{threeparttable}
}
\end{table*}

Recent multimodal large language models (MLLMs) \cite{gpt4,chen2024expanding,wu2024deepseek,Qwen2.5-VL,zhang2024unleash,yang2024heie}, excel in logical reasoning and generalization. These pre-trained models effectively address the limitations of specialized small 3D perception models with their strengths in instruction following, logical reasoning, and cross-domain tasks. However, vanilla MLLMs face issues in specific 3D perception tasks, hindering direct application.

1. \textbf{Weak 3D local spatial object perception:} MLLMs like GPT-4o \cite{gpt4} are primarily trained on 2D data, limiting their ability to capture 3D spatial structures. This results in poor 3D perception performance (Fig. \ref{mllmissue} (a)). Furthermore, these models struggle with capturing fine-grained details of distant and small objects, which is crucial for autonomous driving, as shown in Fig. \ref{mllmissue}(a).

2. \textbf{Poor text-based geometric numerical output:} Existing MLLMs output numerical results in textual form (Fig. \ref{mllmissue} (b)), unsuitable for 3D values (e.g., $X$, $Y$, $Z$, dimensions, rotation), leading to precision issues, slow processing, and parsing difficulties. Outputting structured numerical text is error-prone, often resulting in misordering or misformatting.

3. \textbf{Inability to handle camera focal variations:} Inferring depth from 2D images is under-constrained and disturbed by camera focal length variations. As illustrated in Fig. \ref{mllmissue} (c),  when two objects have the same 2D and 3D sizes, neural networks tend to predict the same 3D location. However, these two images were captured by cameras with different focal lengths, leading to substantial differences in the actual spatial positions.

To overcome the issues of specialized small models and MLLMs, we propose \textbf{\abb}, an M\textbf{LLM}-based \textbf{I}mage \textbf{3D} perception model. By fine-tuning a pre-trained MLLM with LoRA \cite{lora} and proposing a 3D-friendly structure in the image encoder and token decoder, we overcome the limitations of vanilla MLLM, achieving robust 3D perception capabilities.

\begin{figure*}[!t]
\centering
\includegraphics[width=\linewidth]{Figures/threeissue_02122158.pdf}
\caption{\textbf{Three issues of vanilla MLLMs in 3D perception tasks:} (a) Weak 3D local spatial object perception: MLLMs struggle with accurate 3D object localization due to poor spatial understanding, especially for distant or small objects. (b) Poor text-based geometric numerical output: Current models output 3D values in text, which is slow and error-prone. Our approach utilizes a learnable 3D Query token with 3D heads to regress geometric values, improving accuracy significantly. (c) Inability to handle camera focal variations: Distinguishing changes in camera focal length from a single 2D image is hard. This leads to incorrect depth predictions for similarly sized objects captured at different focal lengths.}
\label{mllmissue}
\end{figure*}

In the image encoder, to address MLLMs' weak 3D local spatial object perception, we introduce Spatial-Enhanced Local Feature Mining to extract local 3D image features while reducing token count. We utilize CNNs and depth predictors to enhance local 3D features in high-resolution images. We employ ViT \cite{clip} for global feature extraction with fewer token counts in low-resolution images.  Spatial-enhanced cross-branch attention is then employed to integrate local and global 3D features, further minimizing token number.

In the LLM component, to overcome poor text-based geometric numerical output, we propose the 3D Query Token-Derived Info Decoding method. We utilize a learnable 3D Query token and 3D heads to regress 3D attributes. This learnable 3D Query token can adaptively extract 3D features from images and text. We then use the hidden features from the final LLM layer of the 3D Query to precisely regress the 3D attribute, as illustrated in Fig. \ref{mllmissue}(b).

For 3D box outputting, to address MLLMs' inability to handle variations in camera focal length, we do not solely rely on focal length-invisible black-box neural networks. Instead, we combine black-box networks with white-box projection methods. We introduce Geometry Projection-Based 3D Reasoning, integrating camera parameters into geometric projection to mitigate the impact of varying camera focal lengths on 3D perception.

A comparison of our method with existing specialized small models and large models is shown in Table \ref{introcampare}.

Furthermore, an appropriate dataset is crucial for fine-tuning MLLMs. Existing image-based 3D perception datasets focus on object detection and lack fine-grained caption and question-answer data.
The Mono3DRefer \cite{mono3dvg} dataset also has significant issues. It includes 3D perception results in object caption inputs, which undermines the evaluation of real 3D perception capabilities.

Therefore, we further developed \textbf{\abbdata}: an \textbf{I}mage-based \textbf{3D} \textbf{G}rounding dataset. The \abbdata~dataset provides precise descriptions of objects within images, including detailed appearance and location information.
This enables the 3D grounding task to be effectively performed. Furthermore, our \abbdata~dataset includes annotations for Visual Question Answering (VQA) instructions, allowing the assessment of a model's logical reasoning capabilities and accommodating users' personalized input requirements.

In summary, our contributions are as follows:

1. We are the first to use parameter-efficient fine-tuning to adapt an MLLM for image-based 3D perception, overcoming specialized models' limitations in instruction following, logical reasoning, and open scenario generalization. Furthermore, we identify and resolve three key issues of vanilla MLLMs for 3D perception.

2. To address the issue of weak 3D local spatial object perception, we propose a spatial-enhanced local feature mining approach. This method integrates features extracted by ViT, CNN, and depth predictor while employing the spatial-enhanced cross-branch attention to effectively capture local spatial features of objects.

3. To overcome the problem of poor text-based geometric numerical output, we propose 3D Query token-derived info decoding, which uses a single learnable 3D Query to efficiently extract 3D features within the LLM and regress the 3D values accurately.

4. To mitigate the inability of MLLMs to handle camera focal variations, we propose geometry projection-based 3D reasoning, which combines black-box neural networks with white-box projection methods. By integrating camera parameters, we reduce the significant impact of varying camera focal lengths on 3D perception.

5. We construct the \abbdata, an image-based 3D perception dataset designed to effectively assess a model's 3D grounding and question-answering capabilities.
Extensive experiments demonstrate that our approach achieves state-of-the-art performance on various datasets, surpassing other methods by a large margin.

\section{Related Works}
\label{relatedworks}

\subsection{Multimodal Large Language Models}
GPT-4V \cite{gpt4} integrates image inputs into its language model, demonstrating advanced vision-language multimodal capabilities. GPT-4o \cite{gpt4o} further enhances capabilities across language, audio, images, and video, maintaining rapid response times.
Advancements in open-source MLLM \cite{llava} have also significantly contributed to the development of the field. DeepSeek-VL2~\cite{wu2024deepseek} is an advanced mixture of experts (MoE) model that utilizes dynamic tiling strategies and a MOE language model with multi-head latent attention, enhancing the processing efficiency of high-resolution visual inputs and textual data. InternVL2.5~\cite{chen2024expanding}, building on the InternVL 2.0 architecture, systematically investigates factors in multimodal models, including the impact of visual encoders, language models, dataset size, and inference time on performance. It employs a progressive scaling strategy, initially training the visual encoder with a smaller language model, and then leveraging weight-sharing to transfer this to a larger model, thereby reducing computational demands. Qwen2.5-VL~\cite{Qwen2.5-VL} acts as a visual agent capable of automating computer and mobile operations while understanding lengthy videos and identifying events at specific timestamps. It achieves hierarchical object localization and outputs standardized in JSON format using diverse rectangular boxes and points.

These models excel in 2D tasks with large datasets but lack 3D data training, resulting in weak 3D spatial
perception capabilities.

Recently, the works CubeLLM \cite{cubellm} and EMMA \cite{emma} on arXiv have explored image-based 3D perception. They output geometric coordinates in text format, which affects accuracy and speed. Both models overlook the impact of camera focal length on 3D perception, which can degrade performance with varying focal lengths.
Furthermore, CubeLLM involved substantial resources for pre-training on 2D and 3D alignment, utilizing 9.6 million images and 40.9 million dialogues. The training required 64 A100 GPUs, and its code, model, or dataset has yet to be released. Similarly, EMMA also employed extensive resources for pre-training and has not opened its code.
In contrast, our model, \abb, achieves a 3D-friendly MLLM structure using LoRA \cite{lora}, requiring only two A100 GPUs. This parameter-efficient fine-tuning approach significantly reduces costs and enhances flexibility.

\begin{figure*}[htbp]
\centering
\includegraphics[width=\linewidth]{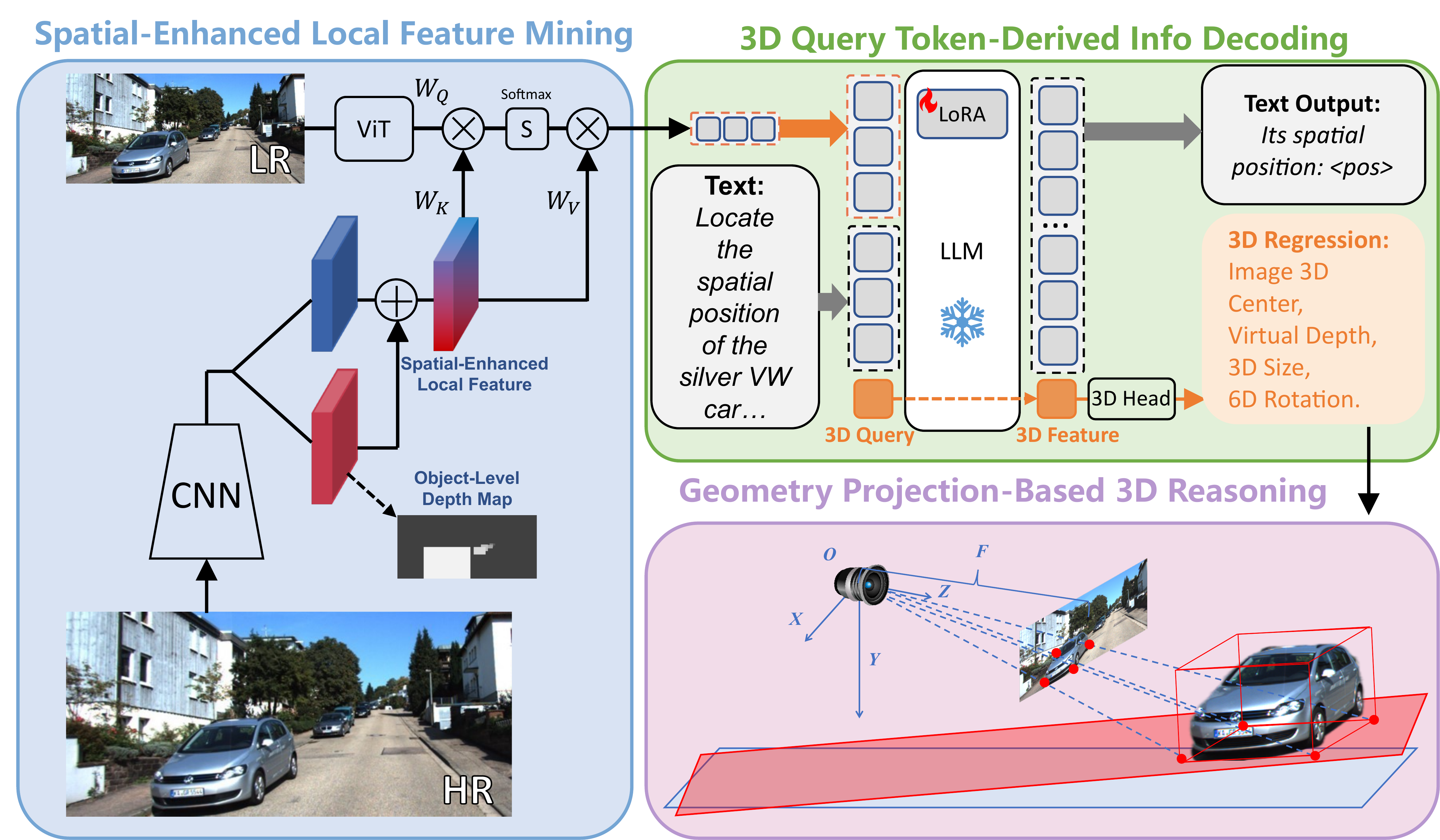}
\caption{\textbf{Framework of the \abb.} (1) The image encoder utilizes Spatial-Enhanced Local Feature Mining, employing a CNN and depth predictor to extract local spatial enhanced features from high-resolution (HR) images. A ViT extracts global features with fewer tokens from low-resolution (LR) images, while spatial-enhanced cross-branch attention efficiently retrieves object spatial features and reduces the token count.
(2) In the LLM, we propose 3D Query Token-Derived Info Decoding. We utilize a learnable 3D Query token to extract 3D features and employ 3D heads to regress the geometric attributes precisely.
(3) To derive the 3D box of the object, we introduce geometry projection-based 3D Reasoning. Rather than using focal length-invisible black-box methods, we combine the network and geometric projection for 3D spatial reasoning, alleviating the errors introduced by varying camera focal lengths in 3D perception.}
\label{pipeline}
\end{figure*}

\subsection{3D Perception from a Single Image}
Research on 3D perception from a single image mainly focuses on monocular 3D object detection. Mono3D \cite{DBLP:conf/nips/ChenKZBMFU15} employed geometric priors and a region proposal network for 3D bounding box (bbox) estimation but was limited by handcrafted features. Deep3DBox \cite{DBLP:conf/cvpr/MousavianAFK17} integrated 2D detection with 3D pose estimation, enhancing localization accuracy through appearance and geometric cues. MonoGRNet \cite{DBLP:conf/aaai/QinWL19} improved precision using graph-based reasoning. RTM3D \cite{DBLP:conf/eccv/LiZLC20} offered an end-to-end framework with advanced feature extraction and attention mechanisms.

Recent advances include MonoFlex \cite{monoflex}, utilizing key points and uncertainty ensembles for better depth estimation, and MonoRCNN \cite{monorcnn}, leveraging geometric data for depth estimation. GUPNet \cite{gupnet} employed 2D and 3D height projection with uncertainty loss for precision, while Gpro3D \cite{gpro3d} improved spatial predictions using ground plane priors \cite{yang2022ground}.

In 2024, Mono3DVG \cite{mono3dvg} was introduced for 3D grounding using text, featuring BERT \cite{bert} and CNN \cite{resnet} for feature extraction. However, due to capacity and pre-training constraints, the model could only process direct descriptions, lacking reasoning and adaptability to new classes, thus limiting generalization.

The Mono3DRefer dataset \cite{mono3dvg}  involves 3D perception results in descriptive inputs, undermining the evaluation of image-based 3D perception.

\section{Methodology}
\label{methodology}

\subsection{Overview}
The architecture of our method is illustrated in Fig. \ref{pipeline}. 
We fine-tune a pre-trained MLLM to empower it with image-based 3D grounding capabilities. 
In the image encoder component, to address the weak 3D local spatial object perception issues of MLLMs, we introduce Spatial-Enhanced Local Feature Mining, enhancing the image encoder's ability to extract local-global spatial features and reducing the token count, as detailed in Section \ref{Spatial-Enhanced Local Feature Mining}.

In the LLM component, to overcome poor text-based geometric numerical output, we propose 3D Query Token-Derived Info Decoding. This method addresses the challenges associated with slow speed, low accuracy, and parsing difficulties encountered when outputting 3D coordinates in text format. By using a single learnable 3D Query token combined with 3D heads regression, we can accurately regress the 3D attributes of objects, as elaborated in Section \ref{3D Query token-derived info decoding}.

To obtain the 3D bounding box (bbox) and address MLLMs' inability to handle variations in camera focal lengths, we introduce geometry projection-based 3D Reasoning. Rather than relying solely on focal length-invisible black-box 3D reasoning methods, we appropriately utilize camera intrinsic parameters. By combining neural networks with geometric projection, our method effectively mitigates the significant errors in 3D perception caused by different camera intrinsic parameters, as detailed in Section \ref{geometry projection-based 3D Reasoning}.

Additionally, in Section \ref{IG3D: Image based 3D Grounding Dataset}, we introduce the \abbdata~dataset. The \abbdata~dataset provides precise descriptions of objects in images, including detailed appearance and location, 
facilitating the 3D grounding task. Moreover, our \abbdata~dataset includes annotations for Visual Question Answering (VQA) instructions, assessing the model's logical reasoning capabilities, and catering to personalized user input requirements.

\subsection{Spatial-Enhanced Local Feature Mining}
\label{Spatial-Enhanced Local Feature Mining}

Image encoders are crucial in multimodal large language models (MLLMs), but face challenges with spatial and local recognition:
(1) MLLMs excel in semantics from 2D data but lack spatial and geometric precision needed for 3D tasks.
(2) Encoders often miss local details \cite{nextchat}, affecting critical tasks like autonomous driving.

On the other hand, Inputting high-resolution images directly into encoders is impractical due to excessive tokens exceeding LLM limits and significantly hindering inference speed.

We propose the Spatial-Enhanced Local Feature Mining algorithm. CNNs and depth predictors enhance local feature extraction in high-resolution images. ViT is used in low-resolution to extract global scene information with fewer tokens. Finally, spatial-enhanced cross-branch attention integrates local and global features while reducing token count.

Unlike the self-attention \cite{DBLP:conf/nips/VaswaniSPUJGKP17} mechanism in ViT \cite{vit}, convolutional layers excel at extracting local features and improving small object identification \cite{li2021localvit}.
Local-enhanced features $F_{\text{local}}$ from ConvNeXt \cite{convnext} are split into two branches: the image RGB feature branch and the spatial depth feature branch. This yields spatial features $F_{\text{spatial}}$ and local RGB features $F_{\text{rgb}}$:
\begin{align}
& F_{\text{spatial}} = Conv_{\text{spatial}}(F_{\text{local}}) \quad F_{\text{rgb}} = Conv_{\text{rgb}}(F_{\text{local}})
\end{align}

We then predict the object-level depth map \cite{DBLP:conf/cvpr/HuangWSH22} using spatial features:
\begin{equation}
    M_{\text{depth}} = Conv_{\text{depth}}(F_{\text{spatial}})
\end{equation}

The spatial depth branch extracts object-level depth features and strengthens the local feature extraction. We use L1 loss and object-level depth ground truth \cite{monodetr} for supervising the depth map.

Image RGB features and spatial depth features are combined to form the Spatial-Enhanced Local Feature \( F_{\text{spatial-local}} \):
\begin{equation}
    F_{\text{spatial-local}} = F_{\text{spatial}} + F_{\text{rgb}}
\end{equation}

We use global feature tokens \( T_{\text{vit}} \) from ViT to adaptively mine the Spatial-Enhanced Local Feature $F_{\text{spatial-local}}$ from the CNN. This process ensures efficient input for the large language model with fewer yet comprehensive tokens. And the spatial-enhanced cross-branch attention employs \( T_{\text{vit}} \) as the query and \( F_{\text{spatial-local}} \) as both key and value:
\begin{align}
   & Q = T_{\text{vit}} \times W_Q  \quad K = F_{\text{spatial-local}} \times W_K \\
   & V = F_{\text{spatial-local}} \times W_V 
    \quad T = Softmax(Q\ K^T / \sqrt{d_k})\ V
\end{align}

We use \( T \) as input for the LLM. \( T \) integrates the advantages of CNN and VIT, enhancing local spatial information extraction with a relatively small number of tokens.

\subsection{3D Query Token-Derived Info Decoding}
\label{3D Query token-derived info decoding}

When MLLMs handle visual perception tasks, they typically output coordinates in the form of text tokens 
\cite{cogvlm,visionllm} or discrete coordinate bin~\cite{DBLP:journals/corr/abs-2306-14824,DBLP:conf/icml/WangYMLBLMZZY22}. To accomplish 3D grounding tasks, a straightforward approach would be to output the object's 3D spatial coordinates in text format, including location, dimension, and rotation.
However, this text-based output method has notable challenges:

1. Low Speed: LLMs like LLaMa \cite{llama} treat each digit as a separate token. For instance, 52.3 uses four tokens. 3D detection outputs (coordinates, dimensions, Euler angles) can require 40-50 tokens (Fig. \ref{mllmissue} (b)), slowing down sequential token generation.

2. Poor Accuracy: LLMs often encounter significant errors when outputting decimal coordinates as text. For example, they generally struggle to accurately interpret the mathematical significance of pitch, roll, and yaw Euler angles in object rotation. Outputting three Euler angles as text poses a significant
challenge for LLMs.

3. Parsing Complexity: With nine degrees of freedom, 3D detection (coordinates, dimensions, angles) often results in LLMs producing incorrect or non-standard outputs, complicating parsing.

Specifically, in the input tokens of the large language model, in addition to the image and text tokens, we introduce a 3D Query token. The 3D Query is a set of learnable parameters that have the same dimension as the hidden layer features of the LLM. 
The function of 3D Query is similar to the query mechanism in DETR \cite{detr}. 
Through adaptive learning, the 3D Query can effectively extract image and text 3D information in the self-attention of the LLM. Only one 3D Query token is needed for this task. 

In LLM, the 3D Query token generates many feature layers, with the final linear layer used to predict the next token for LLM. We select this final LLM layer's hidden state of the 3D Query token as the 3D feature $F_{3D}$.

Additionally, to determine when to use the 3D Query token, we introduce a special \texttt{<pos>} token. The \texttt{<pos>} token is placed before the 3D Query in the sequence. When the \texttt{<pos>} token is detected in the LLM's output, the subsequent next input token is replaced with the learnable 3D Query token. 

Instead of relying on LLM's text output for spatial positioning, we use a regression head. Specifically, with the 3D Feature $F_{3D}$, we employ MLP to regress the object's 3D center projection on the image $p_{\text{img}}$, depth $d_{\text{v}}$, 3D size (length $L$, width $W$, height $H$), and rotation angles.

For the object's center, rather than the 2D box center, we use the projection point of the object's 3D center onto the image, which is more suitable for the subsequent geometric inverse projection process. We use an MLP to predicts the 3D center projection \( p = (u, v) \):
\begin{align}
    & u, v = \text{MLP}_{uv}(F_{3D})
\end{align}

For the 3D size of objects in 3D space: length \( L \), width \( W \), and height \( H \), and the depth \( d_v \), we also use MLP to predict these geometry attributes:
\begin{align}
    L,W,H = \text{MLP}_{LWH}(F_{3D}) \quad d_v = \text{MLP}_{d}(F_{3D}) 
\end{align}

Existing works \cite{cubellm} often predict object rotation using Euler angles, but Euler angles have limitations: 1. Discontinuity: Euler angles can encounter singularities (gimbal lock), causing numerical instability.
2. Non-Uniqueness: Multiple Euler representations for a single rotation hinder model convergence.
3. Complex Loss: Accounting for angle periodicity and ambiguity complicates loss design.
4. Asymmetry: Differing component ranges affect model sensitivity and accuracy.

Additionally, quaternions are also challenging for networks due to discontinuities \cite{DBLP:conf/cvpr/ZhouBLYL19}.

To address these issues, we predict the 6D allocentric rotation~\cite{DBLP:conf/cvpr/ZhouBLYL19}, which is continuous in 6D space and more suitable for learning.
Specifically, we use an MLP to predict the object's 6D allocentric rotation representation:
\begin{align}
    & \text{Rot} = \text{MLP}_{6D}(F_{3D})
\end{align}

In this section, we employ the learnable 3D Query to extract 3D features and regress the geometric attributes from the LLM. We will derive the 3D bounding box in the following section.

\subsection{Geometry Projection-Based 3D Reasoning}
\label{geometry projection-based 3D Reasoning} 

In order to derive the object's 3D box, we need to obtain the spatial location. However, predicting \(X, Y, Z\) coordinates directly is challenging due to varying camera parameters.
Fig. \ref{mllmissue}(c) shows that when two objects have the same 2D and 3D sizes, neural networks tend to predict the same 3D location across varying focal lengths, leading to significant errors.

Therefore, We introduce a geometry projection-based 3D reasoning approach that estimates the 2D projection of the 3D center and virtual depth instead of predicting \(X, Y, Z\) directly.

Camera intrinsics affect depth prediction. To mitigate this, we assume that all input images are captured by a virtual camera with unified focal length and resolution \cite{omni3d}. We do not regress the actual depth of the object but instead, regress the virtual depth under the virtual camera. Subsequently, we convert this virtual depth back to the actual depth of the object.

For a space point \(P^i\) and its image projection pixel \(p^i\), we assume that in the virtual camera, \(P^v\) projects to \(p^v\). \(P^v\) and \(P^i\) have the same \(X\) and \(Y\) but differ in depth \(Z\). The pixels \(p^v\) and \(p^i\) coincide on the image.\footnote{In this paper, we use uppercase letters to denote 3D points in space and lowercase letters for 2D pixels on images.}

Given the real camera's intrinsic matrix \(\mathbf{K^i}\) and real image width \(w^i\), point \(P^i = (X^i, Y^i, Z^i)\) is projected onto the image pixel \(p^i = (x^i, y^i)\):
\begin{align}
Z^i [x^i, y^i, 1]^\top = \mathbf{K^i}[X^i, Y^i, Z^i]^\top \  \label{pro1}
\\ \mathbf{K^i} = \begin{bmatrix} f_{x}^i & 0 & c_{x}^i \\ 0 & f_{y}^i & c_{y}^i \\ 0 & 0 & 1 \end{bmatrix} \quad \quad \ \label{K1}
\end{align}
where \(f_{x}^i, f_{y}^i, c_{x}^i, c_{y}^i\) are the intrinsics of camera \(C^i\).

From Equations \ref{pro1} and \ref{K1}, we can get:
\begin{equation}
x^i \cdot Z^i = f_{x}^i \cdot X^i + c_{x}^i \cdot Z^i
\label{xiz}
\end{equation}

Assuming the intrinsics of the virtual camera are \(f_{x}^v, f_{y}^v, c_{x}^v, c_{y}^v\) and the virtual image width is \(w^v\). The virtual point \(P^v\) corresponds to \(P^i\) in terms of \(X\) and \(Y\), differing solely in the \(Z\) dimension. Therefore, \(P^v = (X^i, Y^i, Z^v)\). The image projection of \(P^v\) is given by \(p^v = (x^v, y^v)\). Similar to Eq. \ref{xiz}, we could get the projection formula in the virtual camera:
\begin{equation}
x^v \cdot Z^v = f_{x}^v \cdot X^i + c_{x}^v \cdot Z^v
\label{xvzv}
\end{equation}

Projection pixels \(p^v = (x^v, y^v)\) and \(p = (x^i, y^i)\) align in the image. And the principal point $c_{x}^v$ for the virtual camera corresponds to the principal point of the real camera:
\begin{align}
x^v = \frac{x^i}{w^i} \cdot w^v , \quad \quad
c_{x}^v = \frac{c_{x}^i}{w^i} \cdot w^v
\label{xv}
\end{align}

Substituting Eq. \ref{xv} into \ref{xvzv} yields:
\begin{equation}
\frac{x}{w^i} \cdot w^v \cdot Z^v = f_{x}^v \cdot X + \frac{c_{x}^i}{w^i} \cdot w^v \cdot Z^v
\label{xwwvzv}
\end{equation}

Thus, we have:
$
x  = f_{x}^v \cdot \frac{X}{Z^v} \cdot \frac{w^i}{w^v} + c_x
$

With Eq. \ref{xiz}, we get:
\begin{equation}
\left(f_{x}^v \cdot \frac{X}{Z^v} \cdot \frac{w^i}{w^v} + c_x\right) \cdot Z = f_{x}^i \cdot X + c_{x}^i \cdot Z
\label{fxvxzvwwv}
\end{equation}

By simplifying \ref{fxvxzvwwv}, we derive the virtual depth:
\begin{equation}
Z^v = \frac{f_{x}^v}{f_x^i} \cdot \frac{w^i}{w^v} \cdot Z
\label{zvfrac}
\end{equation}
Here, \(f_{x}^i, f_{x}^v\) are the focal lengths and \(w^i, w^v\) are image widths of the real and virtual cameras. \(Z\) is the actual depth. 

Virtual depth \(Z^v = \frac{f_{x}^v}{f_x^i} \cdot \frac{w^i}{w^v} \cdot Z\) allows predictions invariant to camera parameters.

Thus, in Section \ref{3D Query token-derived info decoding}, we regress the virtual depth, represented by \(d_v\) from the MLP, and convert it to actual depth.

We invert Eq. \ref{zvfrac} to convert the predicted virtual depth into the real depth:
\begin{equation}
    Z_1 = d_v \cdot \frac{f_{x}}{f_x^v} \cdot \frac{w^v}{w}
    \label{virtualdepth}
\end{equation}
where \( f_x \) and \( w \) denote the focal length and width of the real camera, while \( f_x^v \) and \( w^v \) refer to those of the virtual camera.

In outdoor autonomous driving scenarios, object depths vary significantly. We adopt a geometric projection constraint in the $Y$ direction to estimate a second independent depth. 
Using projection equations (Eq. \ref{pro1} and \ref{K1}), we could get the second depth:
\(    Z_2 = \frac{H}{h}\cdot f_y  \), where $h$ and $H$ are the objects' 2D and 3D height, respectively.

We then compute the average of \(Z_1\) and \(Z_2\) to enhance depth prediction accuracy:
\begin{equation}
    Z = \frac{Z_1 + Z_2}{2} = \left(d_v \cdot \frac{f_{x}}{f_x^v} \cdot \frac{w^v}{w} + \frac{H}{h}\cdot f_y 
 \right) / 2
 \label{zz1z2}
\end{equation}

Currently, we have obtained the Z-coordinate of the 3D bbox center \(P\). Next, 
using the 3D center image projection point \( p = (u, v) \) predicted in Section \ref{3D Query token-derived info decoding}, along with the depth \(Z\), we can calculate the X and Y coordinates of \(P\). According to the projection formula Eq. \ref{pro1} and \ref{K1}, we could get:
\begin{align}
    X = \frac{Z}{f_x} \cdot (u-c_x), \quad
    Y = \frac{Z}{f_y} \cdot (v-c_y)
\end{align}
Substituting from Eq. \ref{zz1z2}, we derive:
\begin{align}
   & X =  \left(\frac{d_v}{f_x^v} \cdot \frac{w^v}{w} + \frac{H}{h}\cdot \frac{f_y}{f_x} 
 \right) \cdot (u-c_x) / 2   \\
 & Y =  \left(\frac{f_{x}}{f_y} \cdot \frac{d_v}{f_x^v} \cdot \frac{w^v}{w} + \frac{H}{h} 
 \right) \cdot (v-c_y) / 2 \\
& Z = \left(d_v \cdot \frac{f_{x}}{f_x^v} \cdot \frac{w^v}{w} + \frac{H}{h}\cdot f_y 
 \right) / 2
\end{align}

Thus, we express the 3D bbox center \(P = (X, Y, Z)\), with neural network outputs and known parameters.

Additionally, in Section \ref{3D Query token-derived info decoding}, we obtained the dimensions of the 3D bbox: \(L\), \(W\), and \(H\), and the rotation \(Rot\). Therefore, we can obtain the final 3D bbox of the object.

\subsection{IG3D: Image-Based 3D Grounding Dataset}
\label{IG3D: Image based 3D Grounding Dataset}

3D grounding requires data with bounding boxes and object descriptions. Current datasets like KITTI \cite{kitti}, nuScenes \cite{nuscenes}, and Waymo \cite{waymo} offer only categories. 
When an image contains multiple objects of the same category, it is impossible to identify the interested object based solely on the category.
Therefore, we propose the \abbdata~dataset, which offers detailed descriptions to distinguish similar objects and supports visual question answering to personalized queries. 

\begin{figure}[t]
\centering
\includegraphics[width=\linewidth]{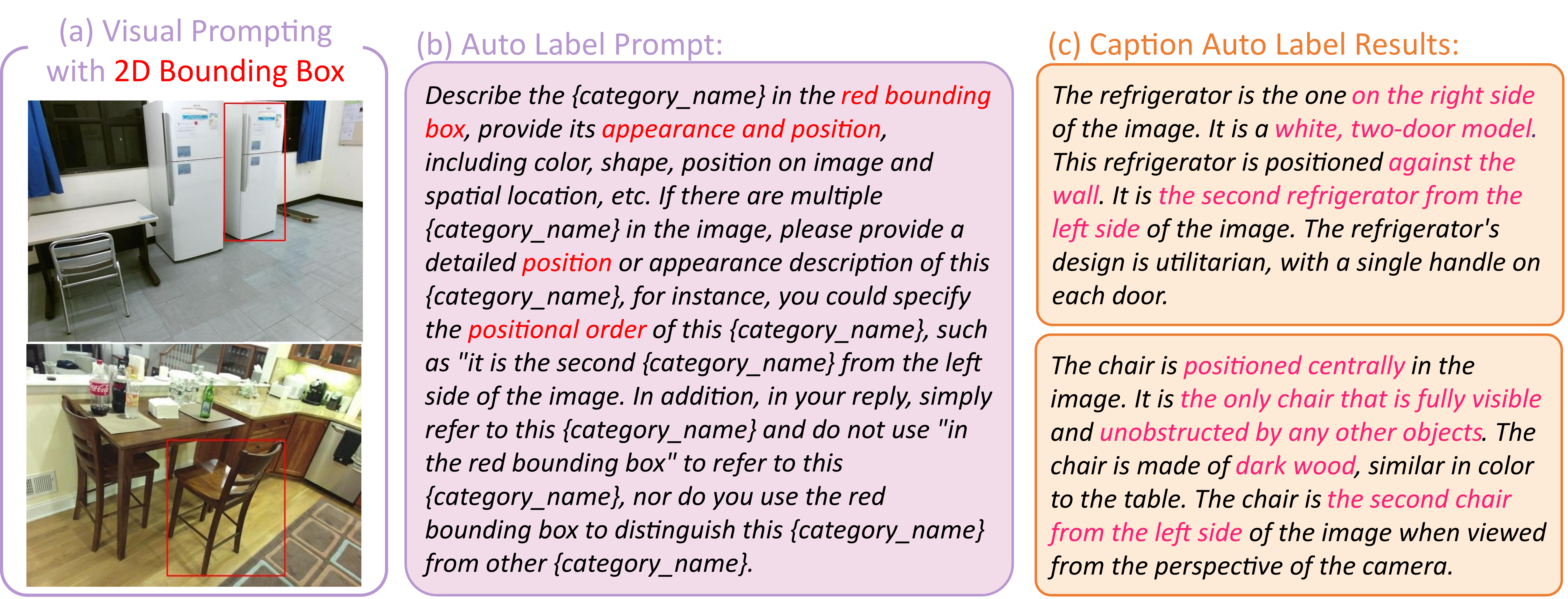}
\caption{\textbf{The auto label process in our \abbdata~dataset.} For each image, we use visual prompting \cite{VisualGPT} to add a 2D box around the target object. A pre-trained MLLM then generates descriptive captions. In this auto label prompt, \{category\_name\} is replaced with the object's category name.}
\label{groundingdataset}
\end{figure}

As shown in Fig.\ref{groundingdataset}, we first employ visual prompting \cite{VisualGPT} to draw the 2D box to highlight the target object in the image. Then, employing a pre-trained MLLM model (Mini-Gemini-34B \cite{mgm}), we produce detailed object descriptions for those objects.

To evaluate Visual Question Answering, we use GPT-4o \cite{gpt4o} to generate specific questions. For example, GPT-4o generated the question: ``I want to read a book, but the light is too dim. Which object should I use?"
Then, the 3D perception model is expected to identify “the lamp” and its 3D box.

We created \abbdata~datasets from SUNRGBD \cite{sunrgbd}, nuScenes \cite{nuscenes}, KITTI \cite{kitti}, and Objectron \cite{objectron}, termed \abbdata-SUNRGBD, \abbdata-nuScenes, \abbdata-KITTI, and \abbdata-Objectron. The \abbdata-SUNRGBD-VQA dataset supports VQA tasks. 
For 3D box ground truth, we use the original annotations from these datasets.
Dataset splits match Omni3D~\cite{omni3d}. We refined the datasets by filtering erroneous annotations. Eight human experts reviewed 30\% dataset annotations to ensure quality.

\section{Experiments}
\label{Experiments}

\begin{table*}[]
\centering
\caption{Comparison of our \abb~with other methods on the \abbdata-SUNRGBD, \abbdata-SUNRGBD-VQA, \abbdata-nuScenes, \abbdata-KITTI, and \abbdata-Objectron datasets.}
\begin{tabular}{@{}llllllll@{}}
\toprule
Dataset & Method & Acc@0.25$\uparrow$ & Acc@0.5$\uparrow$ & DepthError$\downarrow$ & LengthError$\downarrow$ & WidthError$\downarrow$ & HeightError$\downarrow$ \\ \midrule
\multirow{4}{*}{IG3D-SUNRGBD} 
 & TransVG + backproj & 5.6 & 0.4 & 0.88 & 0.57 & 0.84 & 0.59 \\
 & Text3D & 11.5 & 1.7 & 0.45 & 0.20 & 0.34 & 0.21 \\
 & Mono3DVG & 25.2 & 6.8 & 0.53 & 0.14 & 0.26 & 0.16 \\
 & \abb~& \textbf{42.3} & \textbf{11.8} & \textbf{0.32} & \textbf{0.12} & \textbf{0.21} & \textbf{0.12} \\ \midrule
\multirow{4}{*}{IG3D-SUNRGBD-VQA} 
 & TransVG + backproj & 2.2 & 0.2 & 0.97 & 0.71 & 0.93 & 0.77 \\
 & Text3D & 7.8 & 1.0 & 0.56 & 0.24 & 0.45 & 0.29 \\
 & Mono3DVG & 9.8 & 1.4 & 0.63 & 0.21 & 0.41 & 0.27 \\
 & \abb~& \textbf{35.1} & \textbf{8.6} & \textbf{0.36} & \textbf{0.16} & \textbf{0.28} & \textbf{0.17} \\ \midrule
\multirow{4}{*}{IG3D-nuScenes} 
 & TransVG + backproj & 8.6 & 3.5 & 7.51 & 2.28 & 0.77 & 0.82 \\
 & Text3D & 13.7 & 5.2 & 4.25 & 1.75 & 0.28 & 0.27 \\
 & Mono3DVG & 27.5 & 9.8 & 2.80 & 0.55 & 0.19 & 0.21 \\
 & \abb~& \textbf{31.6} & \textbf{13.2} & \textbf{2.19} & \textbf{0.50} & \textbf{0.16} & \textbf{0.17} \\ \midrule
\multirow{4}{*}{IG3D-KITTI} 
 & TransVG + backproj & 2.9 & 0.3 & 8.42 & 1.39 & 0.31 & 0.35 \\
 & Text3D & 5.4 & 0.7 & 4.15 & 0.70 & 0.16 & 0.17 \\
 & Mono3DVG & 27.7 & 7.74 & 2.08 & 0.44 & 0.13 & 0.14 \\
 & \abb~& \textbf{32.4} & \textbf{10.3} & \textbf{1.56} & \textbf{0.34} & \textbf{0.11} & \textbf{0.11} \\ \midrule
\multirow{4}{*}{IG3D-Objectron} 
 & TransVG + backproj & 23.0 & 6.7 & 0.14 & 0.05 & 0.03 & 0.05 \\
 & Text3D & 35.4 & 10.7 & 0.08 & \textbf{0.03} & \textbf{0.02} & 0.04 \\
 & Mono3DVG & 45.5 & 12.4 & 0.09 & \textbf{0.03} & \textbf{0.02} & \textbf{0.03} \\
 & \abb~& \textbf{55.6} & \textbf{18.7} & \textbf{0.05} & \textbf{0.03} & \textbf{0.02} & \textbf{0.03} \\ \bottomrule
\end{tabular}
\label{tab:0captboxfpoif}
\end{table*}

\begin{figure*}[!t]
\centering
\includegraphics[width=\linewidth]{Figures/visvqa_02112154.pdf}
\caption{\textbf{Examples of the 3D VQA of our \abb~in the \abbdata-SUNRGBD-VQA dataset.} Our \abb~is capable of understanding user-input personalized questions, leveraging common knowledge and logical reasoning to identify objects of interest, and returning the corresponding 3D bboxes.}
\label{visvqa}
\end{figure*}

\subsection{Experimental Settings}

\subsubsection{Datasets, Metrics, and Baselines}
Experiments were performed on various datasets. SUNRGBD \cite{sunrgbd} and Objectron \cite{objectron} focus on indoor scenes with more categories. NuScenes \cite{nuscenes}, KITTI \cite{kitti}, and Mono3DRefer \cite{mono3dvg} are outdoor autonomous driving datasets.

We follow the metrics in Mono3DVG \cite{mono3dvg}: ``Acc@0.25" and ``Acc@0.5" for IoU thresholds of 25\% and 50\%, respectively. 
``DepthError", ``LengthError", ``WidthError", and ``HeightError" assess depth and dimension errors in meters.

``Text3D" refers to the results obtained from the baseline Mini-Gemini-7B~\cite{mgm}, which outputs 3D bbox in the form of text. ``TransVG+backproj" follows the baseline from Mono3DVG~\cite{mono3dvg}, using 2D vision grounding combined with back-projection to adapt the results to 3D. Mono3DVG is currently the SOTA image-based 3D grounding method.

\subsubsection{More Implement Details}

We perform parameter-efficient fine-tuning using LoRA~\cite{lora} to fine-tune Mini-Gemini-7B~\cite{mgm}. The rank of LoRA is set to 64, and alpha is set to 16. For the LLM, we choose the Vicuna-7B version~\cite{chiang2023vicuna}. 
We use the L1 loss for the 3D regression heads. We follow the fine-tuning hyper-parameters set by Mini-Gemini, using the AdamW optimizer and the learning rate 2e-5. The batch size is 4, and gradient accumulation steps are set to 4. The fine-tuning process is conducted with two NVIDIA A100 GPUs.

\subsection{Comparison with Other Methods}
\subsubsection{3D Grounding and 3D VQA Results}

As shown in Table~\ref{tab:0captboxfpoif}, our method achieves SOTA performance. The precision of ``TransVG+backproj" is very low, highlighting the difficulty of 3D grounding, as it requires depth and size estimation for bounding boxes, which is challenging from single images.
``Text3D" underperforms due to directly outputting the 3D numerical value in text form. 
Mono3DVG \cite{mono3dvg} performs well in outdoor scenarios but struggles in indoor scenes like SUNRGBD. Our method exceeds Mono3DVG in various datasets, overcoming its yaw angle prediction limitation through 6D allocentric rotation.

The \abbdata-SUNRGBD-VQA dataset involves question-answering and complex 3D reasoning tasks, evaluating the general abilities of models, which are crucial in fields like robotics and embodied intelligence. TransVG+backproj and Mono3DVG are small specialized models and cannot perform reasoning. Our method, utilizing the general capabilities of LLMs, shows strong performance.
Our \abb~achieves state-of-the-art performance in all datasets, surpassing other methods by a significant margin.

\begin{figure*}[!t]
\centering
\includegraphics[width=\linewidth]{Figures/vissunnus_02112155.pdf}
\caption{\textbf{The 3D grounding visualizations of our \abb~and Mono3DVG \cite{mono3dvg} in the SUNRGBD and nuScenes dataset.} When users provide an image and a caption describing the object of interest, the Mono3DVG method lacks sufficient understanding and reasoning abilities for natural language, often misidentifying the object and producing inaccurate 3D bounding boxes. Our approach, however, generates precise 3D bounding boxes for the specified objects.}
\label{vissunnus}
\end{figure*}

\begin{figure}[!t]
\centering
\includegraphics[width=\linewidth]{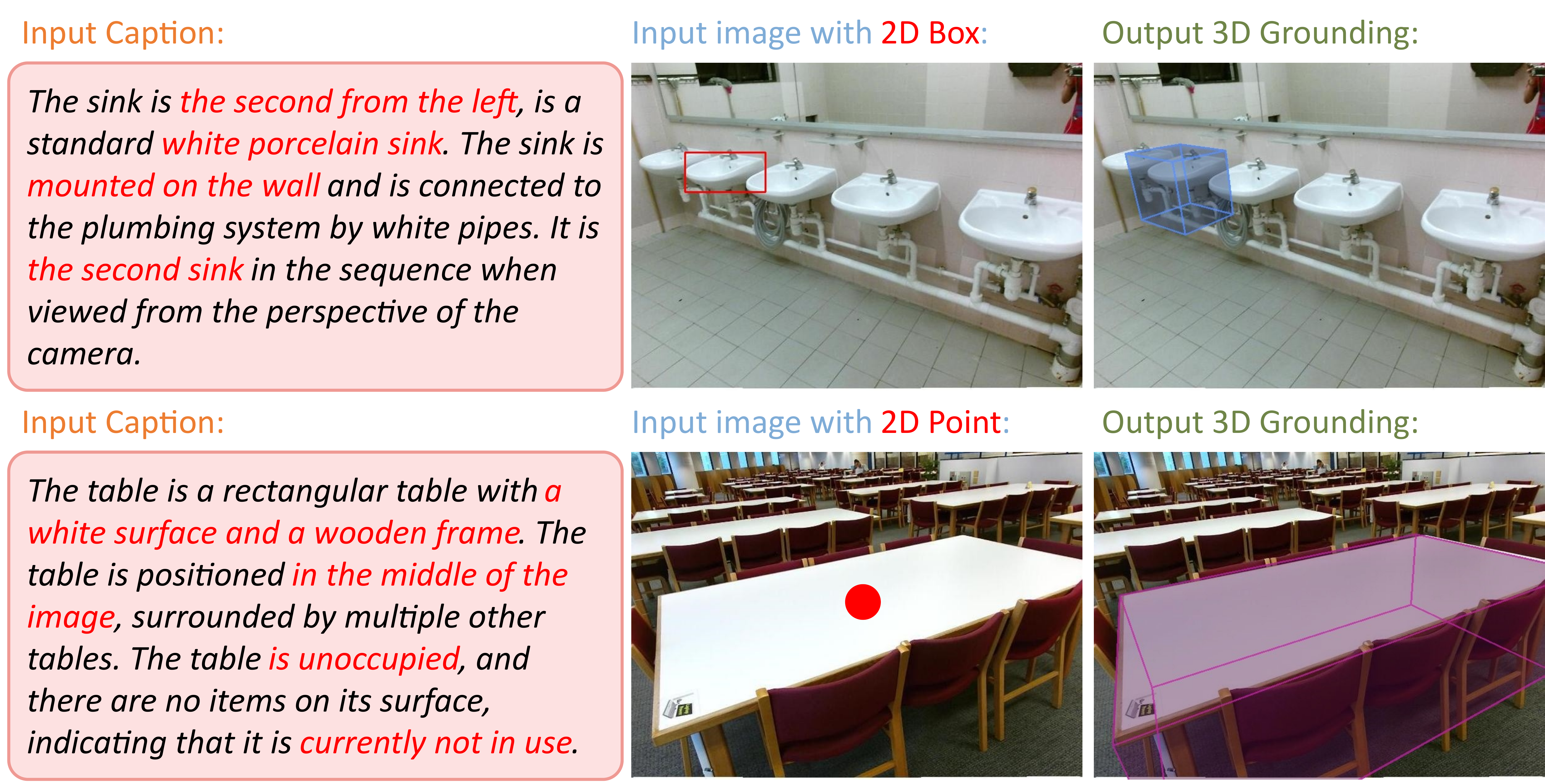}
\caption{Our \abb~can accept various input prompts, such as a caption with a 2D bbox or a caption with a 2D point, and output the corresponding 3D bbox, thereby accommodating various user input forms.}
\label{visboxpoint}
\end{figure}
\subsubsection{Open Vocabulary 3D Grounding}

\begin{table*}[]
\centering
\caption{In the open vocabulary 3D grounding task, results comparison of our \abb~with other methods on the \abbdata-SUNRGBD, \abbdata-nuScenes, \abbdata-KITTI, and \abbdata-Objectron datasets.}
\begin{tabular}{@{}llllllll@{}}
\toprule
Dataset & Method & Acc@0.25$\uparrow$ & Acc@0.5$\uparrow$ & DepthError$\downarrow$ & LengthError$\downarrow$ & WidthError$\downarrow$ & HeightError$\downarrow$ \\ \midrule
\multirow{3}{*}{\abbdata-SUNRGBD} & Text3D & 8.6 & 1.5 & 0.52 & 0.31 & 0.47 & 0.24 \\
 & Mono3DVG & 19.4 & 2.4 & 0.54 & 0.21 & 0.40 & 0.18 \\
 & \abb~& \textbf{40.1} & \textbf{4.4} & \textbf{0.38} & \textbf{0.19} & \textbf{0.29} & \textbf{0.11} \\ \midrule
\multirow{3}{*}{\abbdata-nuScenes} & Text3D & 8.2 & 1.9 & 5.50 & 4.81 & 0.47 & 0.75 \\
 & Mono3DVG & 10.5 & 2.2 & 5.42 & 3.45 & 0.58 & 0.93 \\
 & \abb~& \textbf{30.5} & \textbf{7.5} & \textbf{3.11} & \textbf{2.56} & \textbf{0.36} & \textbf{0.57} \\ \midrule
\multirow{3}{*}{\abbdata-KITTI} & Text3D & 3.1 & 0.3 & 5.30 & 3.88 & 0.37 & 0.62 \\
 & Mono3DVG & 2.2 & 0.2 & 5.97 & 1.88 & 0.46 & 0.51 \\
 & \abb~& \textbf{30.1} & \textbf{8.1} & \textbf{1.84} & \textbf{0.84} & \textbf{0.17} & \textbf{0.26} \\ \midrule
\multirow{3}{*}{\abbdata-Objectron} & Text3D & 20.5 & 6.0 & 0.08 & 0.04 & \textbf{0.03} & \textbf{0.06} \\
 & Mono3DVG & 6.4 & 0.1 & 0.35 & 0.07 & 0.15 & 0.16 \\
 & \abb~& \textbf{32.8} & \textbf{7.9} & \textbf{0.07} & \textbf{0.03} & \textbf{0.03} & \textbf{0.06} \\ \bottomrule
\end{tabular}
\label{openvoc}
\end{table*}

The real world comprises a vast number of categories. Current methods mainly evaluate models on trained categories, overlooking open scenarios.
We assessed zero-shot open-vocabulary 3D grounding by splitting datasets into 80\% base classes and 20\% novel classes. Models trained on base classes were evaluated on novel ones, as summarized in Table \ref{openvoc}.
Table \ref{openvoc} shows our model significantly outperforms others. Models like Mono3DVG perform well on base classes but falter on novel ones.
MLLMs excel in open vocabulary tasks. Unlike small models, which cannot estimate the accurate dimensions of new class objects, large models possess common knowledge and world knowledge, leading to SOTA performance in open scenarios.

\subsubsection{Domain Generalization for the 3D Grounding}

\begin{table*}[]
\centering
\begin{threeparttable}
\caption{In the domain generalization for the 3D Grounding task, results comparison of our \abb~with other methods. nuScenes$\to$KITTI refers to models trained on the nuScenes dataset and tested on the KITTI dataset, while KITTI$\to$nuScenes represents models trained on the KITTI dataset and tested on the nuScenes dataset.}
\begin{tabular}{@{}lllllll@{}}
\toprule
Generalization Setting & Method & Acc@0.25$\uparrow$ & DepthError$\downarrow$ & LengthError$\downarrow$ & WidthError$\downarrow$ & HeightError$\downarrow$ \\ \midrule
\multirow{3}{*}{nuScenes $\to$ KITTI} & Text3D & 5.0 & 4.98 & 0.67 & 0.37 & 0.20 \\
 & Mono3DVG & 16.5 & 2.24 & 0.84 & 0.29 & 0.16 \\
 & \abb~& \textbf{23.1} & \textbf{2.11} & \textbf{0.59} & \textbf{0.20} & \textbf{0.13} \\ \midrule
\multirow{3}{*}{KITTI $\to$ nuScenes} & Text3D & 0.9 & 8.47 & 0.72 & 0.65 & 0.62 \\
 & Mono3DVG & 4.3 & 5.30 & 0.62 & 0.35 & 0.24 \\
 & \abb~& \textbf{20.5} & \textbf{3.32} & \textbf{0.54} & \textbf{0.29} & \textbf{0.22} \\ \bottomrule
\end{tabular}
\label{dgtable}
\end{threeparttable}
\end{table*}

In real-world applications, we encounter not only novel categories but also domain gaps. 
The 3D domain gap is more complex than 2D due to variations in focal lengths and resolutions across datasets, such as lower resolutions in older datasets like KITTI compared to newer ones like nuScenes, significantly affecting 3D localization~\cite{monogdg}.

Table \ref{dgtable} shows domain generalization experiments, using one dataset for training and the other for testing. Notable accuracy drops occur for Text3D and Mono3DVG models due to domain shifts. Our geometry projection-based method interprets images as virtual camera outputs, predicting depth under a unified camera, and performing much better than others.

\subsubsection{Results on Mono3DRefer Dataset}

\begin{table*}[h]
\centering
\caption{
The results on Mono3DRefer Datasets.
}
\label{expmono3drefer}
\begin{tabular}{lcccccccc}
\hline
\multirow{2}{*}{{Method}} &\multirow{2}{*}{{Type}} & \multicolumn{2}{c}{{Unique}}  & \multicolumn{2}{c}{{Multiple}} & \multicolumn{2}{c}{{Overall}} \\ 
 &   & {Acc@0.25$\uparrow$} & {Acc@0.5$\uparrow$} & {Acc@0.25$\uparrow$}  & {Acc@0.5$\uparrow$} & {Acc@0.25$\uparrow$} & {Acc@0.5$\uparrow$}\\ 
 \hline
ZSGNet + backproj  & One-Stage & 9.02 & 0.29 & 16.56  & 2.23  & 15.14 & 1.87\\
FAOA + backproj   & One-Stage  & 11.96  & 2.06 & 13.79 & 2.12 & 13.44 & 2.11 \\
ReSC + backproj   & One-Stage & 11.96   & 0.49 & 23.69 & 3.94 & 21.48 & 3.29\\
TransVG + backproj & Tran.-based & 15.78  & 4.02 & 21.84& 4.16  & 20.70 & 4.14 \\
Mono3DVG-TR & Tran.-based  & 57.65 & 33.04 & 65.92  & 46.85  & 64.36& 44.25 \\ 
\abb~& Tran.-based  & \textbf{60.14} & \textbf{35.91} & \textbf{69.19}  & \textbf{49.20}  & \textbf{67.48}& \textbf{46.70} \\ 
\hline
\end{tabular}
\end{table*}

As shown in Table \ref{expmono3drefer},
to demonstrate the robustness of our model, we evaluated it on the Mono3DRefer dataset \cite{mono3dvg}. We adopt the evaluation metrics from Mono3DVG, where ``unique" refers to images with a single object of a category, and ``multiple" indicates images containing more than one object of the same category.
We follow Mono3DVG's baseline methods.
These 2D grounding baselines, combined with back projection, underperformed on Mono3DRefer, highlighting the complexity of 3D tasks. Our \abb~outperformed all current methods, achieving state-of-the-art accuracy across all metrics, demonstrating our model's strong generalization on diverse datasets.

\subsubsection{Various Types of Input Prompts}

\begin{table*}[h]
\centering
\caption{When the input prompt is changed to caption+2D Box, comparison of our \abb~with other methods on the \abbdata-SUNRGBD, \abbdata-nuScenes, \abbdata-KITTI, and \abbdata-Objectron datasets.}
\label{captboxtpoif}
\begin{tabular}{@{}llllllll@{}}
\toprule
Dataset & Method & Acc@0.25$\uparrow$ & Acc@0.5$\uparrow$ & DepthError$\downarrow$ & LengthError$\downarrow$ & WidthError$\downarrow$ & HeightError$\downarrow$ \\ \midrule
\multirow{3}{*}{IG3D-SUNRGBD} & Text3D & 21.3 & 4.7 & 0.39 & 0.17 & 0.28 & 0.16 \\
 & Mono3DVG & 47.8 & 18.1 & 0.35 & 0.13 & 0.20 & 0.12 \\
 & \abb~& \textbf{56.6} & \textbf{18.6} & \textbf{0.25} & \textbf{0.11} & \textbf{0.18} & \textbf{0.11} \\ \midrule
\multirow{3}{*}{IG3D-nuScenes} & Text3D & 22.4 & 8.9 & 2.16 & 1.19 & 0.26 & 0.19 \\
 & Mono3DVG & 34.1 & 11.5 & 2.31 & 0.46 & 0.17 & 0.15 \\
 & \abb~& \textbf{42.4} & \textbf{17.6} & \textbf{1.45} & \textbf{0.43} & \textbf{0.15} & \textbf{0.14} \\ \midrule
\multirow{3}{*}{IG3D-KITTI} & Text3D & 10.2 & 2.8 & 2.75 & 0.77 & 0.17 & 0.17 \\
 & Mono3DVG & 43.7 & 13.4 & 0.88 & 0.38 & 0.12 & \textbf{0.10} \\
 & \abb~& \textbf{50.1} & \textbf{17.5} & \textbf{0.77} & \textbf{0.34} & \textbf{0.11} & \textbf{0.10} \\ \midrule
\multirow{3}{*}{IG3D-Objectron} & Text3D & 36.7 & 11.5 & 0.08 & \textbf{0.03} & \textbf{0.02} & 0.04 \\
 & Mono3DVG & 53.4 & 16.4 & 0.05 & \textbf{0.03} & \textbf{0.02} & \textbf{0.03} \\
 & \abb~& \textbf{64.4} & \textbf{21.9} & \textbf{0.04} & \textbf{0.03} & \textbf{0.02} & \textbf{0.03} \\ \bottomrule
\end{tabular}
\end{table*}

\begin{table*}[h]
\centering
\caption{When the input prompt is changed to caption+2D Point, comparison of our \abb~with other methods on the \abbdata-SUNRGBD, \abbdata-nuScenes, \abbdata-KITTI, and \abbdata-Objectron datasets.}
\label{captboxfpoit}
\begin{tabular}{@{}llllllll@{}}
\toprule
Dataset & Method & Acc@0.25$\uparrow$ & Acc@0.5$\uparrow$ & DepthError$\downarrow$ & LengthError$\downarrow$ & WidthError$\downarrow$ & HeightError$\downarrow$ \\ \midrule
\multirow{3}{*}{IG3D-SUNRGBD} & Text3D & 15.7 & 2.5 & 0.40 & 0.19 & 0.33 & 0.18 \\
 & Mono3DVG & 42.6 & 11.4 & 0.37 & 0.13 & 0.23 & 0.13 \\
 & \abb~& \textbf{53.7} & \textbf{15.8} & \textbf{0.25} & \textbf{0.12} & \textbf{0.20} & \textbf{0.12} \\ \midrule
\multirow{3}{*}{IG3D-nuScenes} & Text3D & 19.7 & 7.6 & 2.20 & 1.23 & 0.29 & 0.25 \\
 & Mono3DVG & 30.1 & 10.4 & 2.41 & 0.53 & 0.18 & 0.18 \\
 & \abb~& \textbf{37.4} & \textbf{17.4} & \textbf{1.74} & \textbf{0.46} & \textbf{0.17} & \textbf{0.16} \\ \midrule
\multirow{3}{*}{IG3D-KITTI} & Text3D & 8.3 & 2.1 & 3.19 & 0.78 & 0.16 & 0.17 \\
 & Mono3DVG & 31.4 & 9.5 & 1.72 & 0.38 & 0.13 & 0.12 \\
 & \abb~& \textbf{39.0} & \textbf{14.3} & \textbf{1.05} & \textbf{0.34} & \textbf{0.11} & \textbf{0.10} \\ \midrule
\multirow{3}{*}{IG3D-Objectron} & Text3D & 37.9 & 12.6 & 0.07 & \textbf{0.03} & \textbf{0.02} & 0.04 \\
 & Mono3DVG & 49.6 & 15.1 & 0.06 & \textbf{0.03} & \textbf{0.02} & \textbf{0.03} \\
 & \abb~& \textbf{62.8} & \textbf{22.9} & \textbf{0.05} & \textbf{0.03} & \textbf{0.02} & \textbf{0.03} \\ \bottomrule
\end{tabular}
\end{table*}

Table~\ref{captboxtpoif} shows results when the prompt includes the caption and the 2D box (as shown in the upper part of Fig. \ref{visboxpoint}). During input, we draw a 2D box on the image as a visual prompting \cite{VisualGPT} to assist the model in performing 3D grounding. Adding the 2D box to prompt inputs improves accuracy in all datasets compared to caption-only prompts, highlighting the challenge of 3D grounding with only textual cues. When images have multiple overlapping objects within the same category, distinguishing them using text alone is difficult. Utilizing both caption and 2D box as input significantly enhances localization accuracy.

Table~\ref{captboxfpoit} reports results with prompts including text caption and a 2D point (as shown in the lower part of Fig. \ref{visboxpoint}). We draw a 2D point on the input image as a visual prompting \cite{VisualGPT} to assist the model in performing 3D grounding. 
Including a 2D point in the input prompt helps with grounding the object.
In robotics and augmented reality, users may employ various inputs and cues to locate the object of interest, such as a 2D point and textual description.
Our method effectively adapts to these user input types, demonstrating robust and versatile capabilities.

\subsection{Ablation Study}

\subsubsection{Ablation Study on the Spatial-Enhanced Local Feature Mining}

\begin{table}[h]
\centering
\resizebox{1.0\hsize}{!}{
\begin{threeparttable}
\caption{Ablation study on the Spatial-Enhanced Local Feature Mining method using the \abbdata-SUNRGBD dataset. ``SECBA" stands for Spatial-Enhanced Cross-Branch Attention.
 }
\label{imageencoder}
\begin{tabular}{@{}c|ccc|ccc@{}}
\toprule
Exp. & HR Branch & Depth Branch & SECBA & Acc@0.25$\uparrow$ & Acc@0.5$\uparrow$ & DepthError$\downarrow$ \\ \midrule
(a) &  &  &  & 30.7 & 6.1 & 0.73 \\
(b) & \checkmark &  &  & 35.4 & 8.6 & 0.58 \\
(c) & \checkmark & \checkmark &  & 37.4 & 9.3 & 0.42 \\
(d) & \checkmark & \checkmark & \checkmark & \textbf{42.3} & \textbf{11.8} & \textbf{0.32} \\ \bottomrule
\end{tabular}
\end{threeparttable}
}
\end{table}

As shown in Table \ref{imageencoder}, 
the ``HR branch" employs a ConvNeXt \cite{convnext} CNN for high-resolution feature extraction, while a ``depth branch" estimates object-level depth for spatial feature enhancement. ``SECBA" (Spatial-Enhanced Cross-Branch Attention) 
leveraging spatial-enhanced local features from the CNN and the global tokens from the ViT to perform spatial-enhanced cross-branch attention.
Exp. (a) does not use the high-resolution (HR) branch and relies solely on low-resolution ViT tokens, which limited spatial feature extraction, resulting in increased DepthError and decreased accuracy.
Exp. (b) adds a CNN HR branch, enhancing the identification of small and distant objects with high-resolution images. Exp. (c) includes a depth branch in the CNN for better spatial extraction via depth supervision. 
Exp. (b) and (c) naively use max pooling to sum the feature maps extracted by the CNN with the ViT tokens before inputting them into the LLM.
Exp. (d) employs SECBA (Spatial-Enhanced Cross-Branch Attention) to effectively integrate CNN local spatial features with ViT tokens, achieving the highest accuracy.

\subsubsection{Ablation Study on the 3D Query Token-Derived Info Decoding}

\begin{table}[h]
\centering
\caption{Ablation Study on the \abbdata-SUNRGBD dataset to evaluate the decoding methods in the LLM part.}
\label{onetoken}
\begin{tabular}{@{}c|c|ccc@{}}
\toprule
Exp. & Setting & Acc@0.25$\uparrow$ & Acc@0.5$\uparrow$ & DepthError$\downarrow$ \\ \midrule
(a) & Text Output & 11.5 & 1.7 & 0.45 \\
(b) & Text Feature & 21.6 & 4.7 & 0.42 \\
(c) & Position Token & 40.5 & 10.4 & 0.37 \\
(d) & 3D Query Decoder & \textbf{42.3} & \textbf{11.8} & \textbf{0.32} \\ \bottomrule
\end{tabular}
\end{table}

As shown in Table \ref{onetoken}, in Exp. (a), the 3D grounding results are output in text form, leading to parsing difficulties and low accuracy.
Exp. (b) improves accuracy by using the hidden layer features of text tokens with an MLP for regressing 3D values, highlighting the effectiveness of regression over text-based outputs.
In Exp. (c), \texttt{<pos>} token features are used with an MLP, which is effective but less accurate than the 3D Query method.
Exp. (d) presents our comprehensive method, introducing a learnable 3D Query token in the LLM, allowing precise 3D feature extraction with a single 3D Query token.

\subsubsection{Ablation Study on the Geometry Projection-Based 3D Reasoning}

\begin{table}[h]
\centering
\caption{Ablation study on the \abbdata-nuScenes dataset to evaluate the Geometry Projection-Based 3D Reasoning method.}
\label{georeasoning}
\resizebox{1.0\hsize}{!}{
\begin{tabular}{@{}c|ccc|ccc@{}}
\toprule
Exp. & Back Projection & Height Depth & Virtual Depth & Acc@0.25$\uparrow$ & Acc@0.5$\uparrow$ & DepthError$\downarrow$ \\ \midrule
(a) &  &  &  & 19.8 & 8.6 & 3.46 \\
(b) & \checkmark &  &  & 26.9 & 10.4 & 2.94 \\
(c) & \checkmark & \checkmark &  & 28.8 & 12.1 & 2.73 \\
(d) & \checkmark &  & \checkmark & 29.4 & 12.5 & 2.57 \\
(e) & \checkmark & \checkmark & \checkmark & \textbf{31.6} & \textbf{13.2} & \textbf{2.19} \\ \bottomrule
\end{tabular}
}
\end{table}

The nuScenes dataset involves six cameras with varied focal lengths. In Table \ref{georeasoning}, 
``Back projection" uses the predicted 3D center and depth; ``height depth" is via \( Z_2 = \frac{H}{h} \cdot f_y \); and ``Virtual depth" via Eq. \ref{virtualdepth}: \( Z_1 = d_v \cdot \frac{f_x}{f_x^v} \cdot \frac{w^v}{w} \).
Exp. (a) omits back projection, directly regressing \( X \), \( Y \), \( Z \), causing accuracy issues due to the difficulty of predicting these 3D coordinates.
Exp. (b) uses the 3D center, depth, and back projection to derive the 3D location, improving accuracy over (a). Exp. (c) incorporates 2D-3D height depth with \( Z_2 = \frac{H}{h} \cdot f_y \), further reducing depth error.
Exp. (d) introduces virtual depth, benefiting from a uniform virtual camera to address focal length variations across cameras.
Exp. (e) combines 2D-3D height and virtual depth to get the final depth (Eq. \ref{zz1z2}), effectively solving focal length issues, and achieving superior accuracy.

\subsubsection{Experiments on the Rotation Angle Prediction}

\begin{table}[]
\centering
\caption{Experiments on the \abbdata-SUNRGBD dataset to evaluate the rotation angle prediction methods.}
\label{rotationexp}
\begin{tabular}{@{}l|l|ll@{}}
\toprule
Exp. & Rotation Prediction & \multicolumn{1}{c}{Acc@0.25$\uparrow$} & \multicolumn{1}{c}{Acc@0.5$\uparrow$} \\ \midrule
(a) & Euler Angle & 37.1 & 8.8 \\
(b) & 6D Allocentric Rotation & \textbf{42.3} & \textbf{11.8} \\ \bottomrule
\end{tabular}
\end{table}

In Table \ref{rotationexp}, Exp. (a), which uses Euler Angles, and shows substantial errors. Exp. (b) employs 6D allocentric rotation, a continuous representation, which is better suited for neural networks.

\subsection{Visualization Results}

Fig. \ref{visvqa} showcases the model's performance on the \abbdata-SUNRGBD-VQA dataset. Our model effectively answers the questions and locates objects in 3D.

Fig. \ref{vissunnus} illustrates 3D grounding on the indoor and outdoor autonomous driving scenarios of our \abb~and Mono3DVG \cite{mono3dvg}. With image and caption inputs, Mono3DVG \cite{mono3dvg} often struggles with natural language comprehension and reasoning, leading to confusion between objects and incorrect spatial localization. In contrast, our \abb~accurately identifies and localizes the object of interest across diverse environments.

Fig. \ref{visboxpoint} illustrates our \abb~'s capability to accept various input prompts, including captions, 2D boxes, and 2D points, demonstrating its flexibility.

\section{Conclusion}
\label{conclusion}
The demand for 3D perception such as autonomous driving, augmented reality, and robotics is growing rapidly. In this paper, we identified and tackled three major issues faced by MLLMs in 3D perception: (1) Weak 3D local spatial object perception, (2) Poor text-based geometric numerical output, and (3) Inability to handle camera focal variations.

To address these challenges, we proposed \abb, a 3D-friendly MLLMs architecture. For the image encoder, we introduced Spatial-Enhanced Local Feature Mining. In the LLM part, we proposed 3D Query Token-Derived Info Decoding. To obtain the 3D bounding boxes of objects, we proposed Geometry Projection-Based 3D Reasoning. Furthermore, we introduce the \abbdata~dataset, which assesses fine-grained grounding, logical reasoning, and question-answering in 3D perception models.

Extensive experiments on various settings, including 3D grounding and 3D VQA, open-vocabulary 3D grounding, and domain generalization confirm that our method achieves state-of-the-art results, surpassing other methods by a significant margin.
However, our work still has limitations. 
While the MLLM-based 3D perception framework demonstrates strong general capabilities, its inference latency is larger than that of specialized small models. In the future, we aim to improve the reasoning efficiency of the MLLM-based framework.

\bibliographystyle{IEEEtran}
\bibliography{LLMI3D}

\end{document}